\documentclass{amsart}

\usepackage{
amsmath,amsthm,amsfonts,amscd,graphicx,latexsym,amssymb,eucal,mathrsfs
}
\usepackage[all]{xy}

\def\ZR#1{\langle #1\rangle}
\def\longhookrightarrow{\lhook\joinrel\longrightarrow}

\newcommand {\Stab}{{\rm stab}}
\newcommand {\order}{{\rm ord}}
\newcommand {\Star}{{\rm Star}}
\newcommand{\absolute}[1]{\lvert#1\rvert}
\renewcommand{\phi}{\varphi}
\DeclareMathOperator{\Ver}{Vert}

\DeclareMathOperator{\PGL}{PGL}

\newcommand{\mathset}[1]{{\{#1\}}} 

\newtheorem{thm}{Theorem}[section]

\newtheorem{cor}[thm]{Corollary}

\newtheorem{dfn}[thm]{Definition}

\newtheorem{exa}[thm]{Example}

\newtheorem{conv}[thm]{Convention}

\newtheorem{Rem}[thm]{Remark}

\newtheorem*{Satz*}{Satz}

\begin{document}
\title{Degenerating families of dendrograms}

\author{Patrick Erik Bradley}

\date{\today}

\begin{abstract}
Dendrograms used in data analysis are ultrametric spaces, hence objects of nonarchimedean geometry.
It is known that there exist $p$-adic representation of dendrograms. Completed by a point at infinity,
they can be viewed as subtrees of the Bruhat-Tits tree associated to the $p$-adic  projective line.
The implications are that certain moduli spaces known in algebraic geometry are $p$-adic parameter spaces
 of (families of) dendrograms, and stochastic classification can also be handled within this framework.
At the end, we calculate the topology of the hidden part of a dendrogram.  
\end{abstract}

\maketitle

\section{Introduction}
Dendrograms used in data analysis are ultrametric spaces. Hence they are objects of nonarchimedean
geometry, a special instance of which is $p$-adic geometry. 
Murtagh \cite{Murtagh2004a} shows how to associate to a dendrogram a set of $p$-adic representations
of integers. This lies well within the tradition of using ultrametrics in order to describe the hierarchical
ordering  in classification (cf.\ \cite{Murtagh2004} and the references therein). 

However, there is seemingly a problem in the choice of the prime number $p$ for the $p$-adic representation
of dendrograms by the fact that the geometry of the $p$-adic number field $\mathbb{Q}_p$ allows
only at most $p$ maximal subclusters of any given cluster. We will show that this can be overcome
by considering finite field extensions of $\mathbb{Q}_p$, so that the convenient choice $p=2$ becomes feasible
for any dendrogram. This seems to be compliant with the  philosophy of allowing any nonarchimedean complete valued field for describing, coding or computing in data analysis. We acknowledge here our
inspiration by \cite{Murtagh2004}. 

Our point of view is in fact of a geometric nature. For a $p$-adic geometer, a dendrogram
is nothing but the affine $p$-adic line $\mathbb{A}^1$ with $n$ punctures from which a 
certain kind of covering of $\mathbb{A}^1$ can be made
whose intersection graph
 is the tree
 in bijection with the  dendrogram from the point of view of data analysis.
Completing the affine line to the projective line $\mathbb{P}^1$ and then taking an extra puncture $\infty$, 
allows us to see the dendrogram as a subtree of the {\em Bruhat-Tits tree},
which is an important object in the study of $p$-adic algebraic curves.
A first application is in the coding of DNA sequences \cite{DragovichDragovich}, 
which is a special case of $p$-adic methods for
 processing strings over a given alphabet, as 
explained in \cite{Brad2007}, where also new invariants
of time series of dendrograms are developped.

It is an imperative  from the geometric viewpoint to study
families of dendrograms. For these, there exist already parameter spaces.
In fact, it is now the {\em moduli space of genus $0$ curves with $n$ punctures} $M_{0,n}$
from algebraic geometry
which now becomes the central object of interest. Each point of the $p$-adic version of $M_{0,n}$
is a dendrogram with the extra point $\infty$.  It is then a natural consequence that a
stochastic dendrogram is a continuous family of dendrograms together with a probability distribution on it,
or, we can make this now more precise, a map from a $p$-adic set of parameters $S$ to $M_{0,n}$
with a probability distribution on $S$. 
 We will give an idea of $p$-adic spaces
by explaining the {\em Berkovich topology} one has on these. Due to the ultrametric property, 
$p$-adic spaces in a na\"{\i}ve sense are totally disconnected. This problem can be remedied by
introducing extra points which can, in a generalised sense, be viewed as clusters of usual points.

In this framework, collisions of points in their evolution through time can be formally described
by considering the compactification $\bar{M}_{0,n}$ by stable trees of projective lines
which we  call {\em stable dendrograms}. 
Time series of dendrograms, on the other hand, yield (analytic) maps $M_{0,m}\to M_{0,n}$ between
the moduli spaces. 
Further applications of these moduli spaces should be in the study of consensus
of dendrograms.

We end by calculating the topology of the {\em hidden part} of a dendrogram,
i.e.\ the subgraph spanned by vertices corresponding to clusters which do not have singletons as maximal subclusters. This subgraph determines the distribution of the other clusters,
which are ``near the end'' of the
dendrogram.

\medskip
An introduction to $p$-adic numbers is \cite{Gouvea1993}. Algebraic curves 
can be learned with
a minimum amount of technical requirements in \cite{Griffiths1989}. A bird's eye on moduli spaces of curves 
is found in \cite[Appendix: Curves and Their Jacobians]{Mumford1999}. 
A broader introduction to moduli of curves is \cite{HarrisMorrison}. A non-technical introduction to Berkovich spaces and analysis 
on the projective line is contained in 
\cite{Baker2004, BakerRumely}. Those who intend an intensive
study of these subjects might wish to learn more algebraic geometry
which can be found in  
\cite{Mumford1999}. 

\section{Dendrograms and nonarchimedean geometry}

\begin{figure}[h]
\centering
\includegraphics[scale=.3]{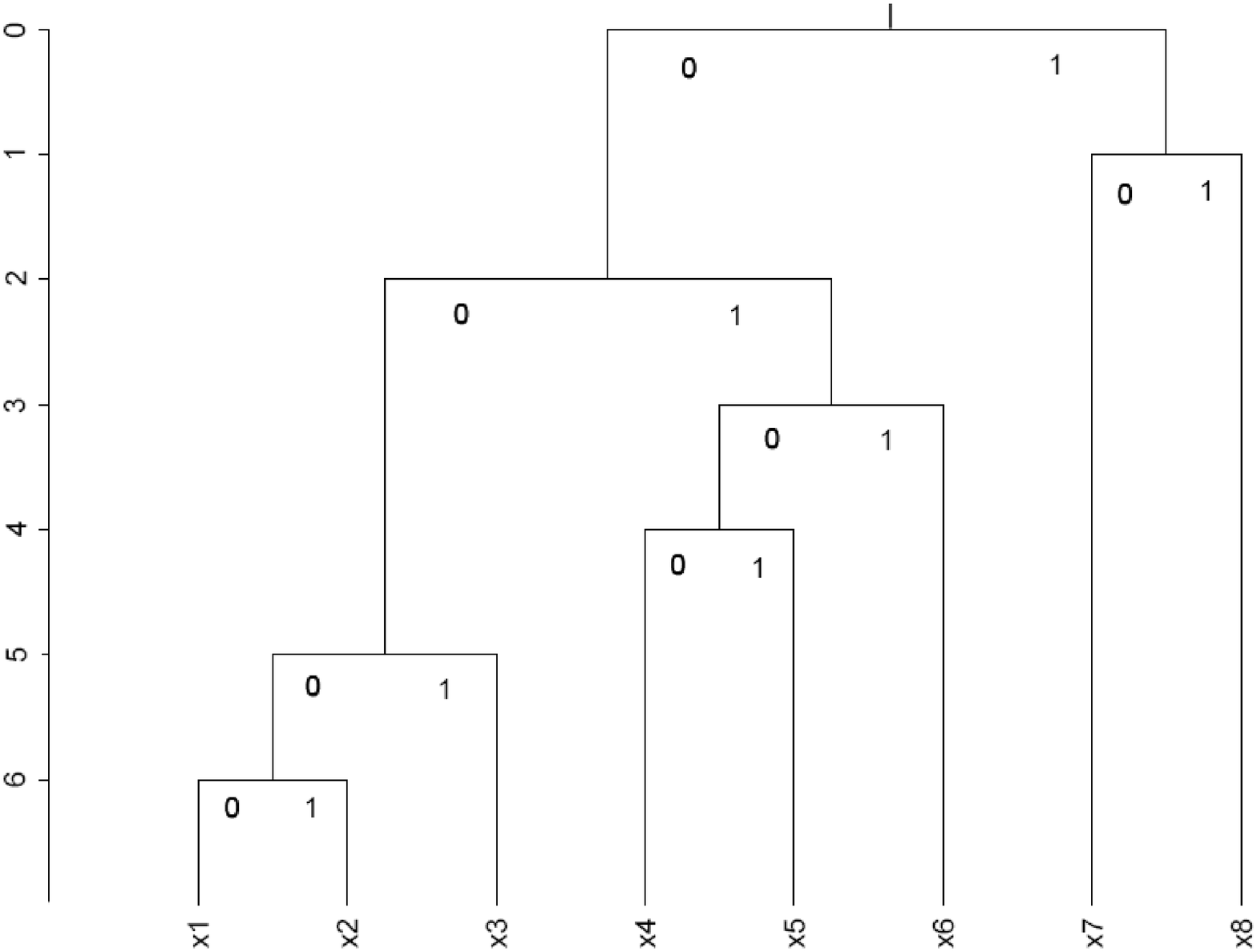}
\caption{A $2$-adic dendrogram.}
\label{dendo}
\end{figure}

Dendrograms are known to be endowed with a nonarchimedean metric, also called an ultrametric, for
which the strict triangle inequality 
$$
d(x,y)\le\max\mathset{d(x,z),d(z,y)}
$$
holds.
Therefore, it is quite tempting to use $p$-adic numbers for their description, and in fact, this has recently been
done \cite{Murtagh2004,Murtagh2004a}. I shall explain this along the example dendrogram
of Figure \ref{dendo}, which is a slight modification of \cite[Fig.\ 1]{Murtagh2004a}. Choose a prime number $p$, and distribute the $p$ numbers $0,\dots,p-1$ across the partitioning of the horizontal
line segments defined by the intersection points with vertical line segments of the dendrogram.
For the top horizontal line segment, one has to introduce one extra vertical line segment
going upwards\footnote{The usefulness of this  extra detail 
will become apparent in the following sections.}, as effected in Figure \ref{dendo}. On going down on a path $\gamma$
from the top vertical line segment
all the way down to one of the points $x_i$, one picks up the numbers $\alpha$ on the traversed horizontal line segments $\ell$ and obtains
$$
x=\sum\limits_\nu\alpha_\nu p^\nu,
$$
where $\nu=\nu(\ell)$ runs through all levels of the horizontal parts $\ell$ of the path $\gamma$.

In our example from Figure \ref{dendo}, we assume $p=2$, and obtain the numbers
$$
\begin{array}{llll}
x_1=0, \quad & x_2=2^6,\quad &x_3=2^5,\quad& x_4=2^2,
\\
x_5=2^2+2^4,\quad &x_6=2^2+2^3,\quad& x_7=2^0,\quad& x_8=2^0+2^1.
\end{array}
$$
Note that these dyadic representations differ from the ones in \cite[\S 2]{Murtagh2004a}.
In any case, each path from the top to a bottom end of the dendrogram corresponds to a $p$-adic power series representation of an integer number.
The choice of the prime $p$ is arbitrary. However, it might seem that the possible number of vertical segments attached to one horizontal line segment allowing  a $p$-adic representation of a dendrogram might be bounded by $p$. But this is not the case.
In fact, one can restrict to the arbitrary choice $p=2$, if one wishes, and can describe  all dendrograms by the
help of a little algebra, as will be seen in the following section.

\section{The Bruhat-Tits tree} \label{Bruhat-Tits-tree}

Let $\mathbb{Q}_p$ be the field of $p$-adic numbers. It is a complete nonarchimedean normed field whose norm
will be denoted by $\absolute{\cdot}_p$. Consider the unit disk
$$
\mathbb{D}=\mathset{x\in\mathbb{Q}_p\mid \absolute{x}_p\le 1}=B_1(0).
$$
It contains the $p$ maximal smaller disks 
$$
B_{\frac{1}{p}}(0),\;B_{\frac{1}{p}}(1),\dots, B_{\frac{1}{p}}(p-1)
$$
corresponding to the residue field $\mathbb{F}_p$ of $\mathbb{Q}_p$. This well known fact is actually a consequence
of the construction from the previous section.

It is useful to consider the $p$-adic projective line $\mathbb{P}(\mathbb{Q}_p)=\mathbb{Q}_p\cup\mathset{\infty}$, in which there is
the maximal disk outside $\mathbb{D}$:
$$
\mathset{x\in\mathbb{P}(\mathbb{Q}_p)\mid\absolute{x}_p\ge p}=B_p(\infty).
$$
Due to the ultrametric topology on the $p$-adic projective line, the ``closure'' of an ``open'' disk
depends somewhat on the choice of a point on its ``boundary'' \cite[\S1.1]{Gerritzen1978}.
Therefore, we make 
\begin{dfn}
Let 
$$
B=\mathset{x\in\mathbb{P}(\mathbb{Q}_p)\mid \absolute{x-a}_p<r} \quad(\text{resp.}\quad B=\mathset{x\in\mathbb{P}(\mathbb{Q}_p)\mid \absolute{x-a}_p>r})
$$ 
for some $a\in\mathbb{Q}_p$ and a $p$-adic value $r=\absolute{\epsilon}_p$, $\epsilon\in\mathbb{Q}_p\setminus\mathset{0}$,
and let $b\in \mathbb{Q}_p$ such that $\absolute{a-b}_p=r$.
The {\em affinoid closure of $B$ with respect to $\infty$} (resp.\ {\em to $b$}) is the disk
$$
\bar{B}=\mathset{z\in\mathbb{P}(\mathbb{Q}_p)\mid\absolute{x-a}_p\le r}\quad (\text{resp.}\quad \bar{B}=\mathset{z\in\mathbb{P}(\mathbb{Q}_p)\mid\absolute{x-b}_p\ge r}).
$$
\end{dfn}

Using the projective line necessitates the introduction of an equivalence relation  on the set
of all disks of $\mathbb{P}(\mathbb{Q}_p)$. Namely, disks $B_1$, $B_2$ are said to be {\em equivalent:} $B_1\sim B_2$,
if either $B_1=B_2$ or the  affinoid closure of $\mathbb{P}(\mathbb{Q}_p)\setminus B_2$ with respect to some point 
$a\in B_2$
equals $B_1$ \cite[\S 1]{Herrlich1980}. One checks that the relation $\sim$ is indeed an equivalence relation.

The {\em Bruhat-Tits tree} $\mathscr{T}_{\mathbb{Q}_p}$ is defined by setting its vertices  to be the equivalence classes of
disks in $\mathbb{P}(\mathbb{Q}_p)$, and its edges are given by maximal inclusion of disks, i.e.\
an edge $e=([B_1],[B_2])$ means that 
$B_1$ is strictly contained in $B_2$, and $B_1$ is a maximal disk with this property, for suitable representative disks.
It is a well known fact that $\mathscr{T}_{\mathbb{Q}_p}$ is indeed a tree. This can be seen directly in this way:
Each class is obviously represented by a unique disk $B$ which is the closure with respect to $\infty\notin B$, 
 and the disks not containing infinity are preordered by inclusion; so $\mathscr{T}_{\mathbb{Q}p}$ is a directed acyclic graph, hence a tree by the ultrametric
 property of $\absolute{\cdot}_p$. 

\begin{figure}
\centering
\includegraphics[scale=.3]{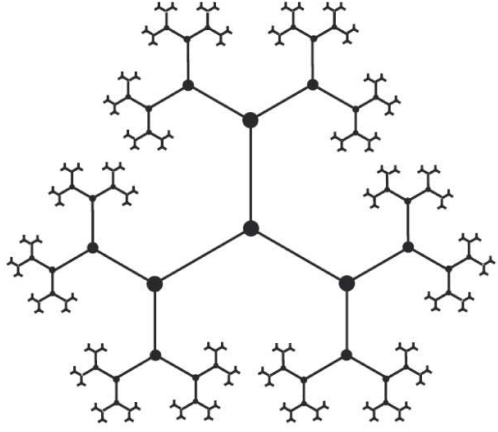}
\caption{The Bruhat-Tits tree for $\mathbb{Q}_2$.} \label{btt2}
\end{figure}

The {\em star}  of a vertex $v$ in $\mathscr{T}_{\mathbb{Q}_p}$, denoted as $\Star_{\mathscr{T}_{\mathbb{Q}_p}}(v)$, consists of all edges emanating from  $v$.
The edges of any star are in one-to-one correspondence with the points of $\mathbb{P}(\mathbb{F}_p)=\mathbb{F}_p\cup\mathset{\infty}$, i.e.\ the $\mathbb{F}_p$-rational points of the projective line over the residue field $\mathbb{F}_p$. Namely, this is true for the vertex $v_\mathbb{D}$ corresponding to the unit disk $\mathbb{D}$,
and the  group of M\"obius transformations acts on $\mathscr{T}_{\mathbb{Q}_p}$ \cite[Bemerkung 5]{Herrlich1980}.
Thus the Bruhat-Tits tree $\mathscr{T}_{\mathbb{Q}_p}$ is a $p+1$-regular locally finite tree. An illustration of $\mathscr{T}_{\mathbb{Q}_2}$ from \cite[Fig.\ 5]{CK2005} is given
in Figure \ref{btt2}.

\medskip
By construction, the tree $\mathscr{T}_{\mathbb{Q}_p}$ is invariant under
transformations of the form $z\mapsto\frac{az+b}{cz+d}$,
with $a,b,c,d\in\mathbb{Q}_p$ such that $ad-bc\neq 0$.
These transformations are called {\em projective linear} or {\em M\"obius transformations}, and form the group $\PGL_2(\mathbb{Q}_p)$. The reason for
invariance under $\PGL_2(\mathbb{Q}_p)$ is the well known fact that M\"obius
transformations take equivalent disks to equivalent disks.

\medskip
As it may happen that a cluster may have more than $p$ maximal subclusters,
it would be convenient to be able to represent such dendrograms without enlarging the prime $p$.
So, let $K\supseteq\mathbb{Q}_p$ be a finite extension field of $\mathbb{Q}_p$. The $p$-adic norm extends,
similarly as in the archimedean case, uniquely
to an ultrametric norm $\absolute{\cdot}_K$ on $K$, and $K$ is complete with respect to $\absolute{\cdot}_K$. 
Such a field $K$ is called a {\em $p$-adic field}.

For a $p$-adic field $K$, there is in a similar manner as for $\mathbb{Q}_p$ a Bruhat-Tits tree $\mathscr{T}_K$. Again $K$ has a finite residue field with $q=p^m$ elements, and $\mathscr{T}_K$ is $q+1$-regular. Therefore, in
practical applications it should be possible to stick to the prime $p=2$ and make finite field extensions,
if there are clusters with more than $2$ children clusters. Again, $\PGL_2(K)$
respects the symmetries of the hierarchical structure of 
the Bruhat-Tits tree,
i.e.\ $\mathscr{T}_K$ is invariant under projective linear transformations
defined over $K$.

\bigskip
For convenience, we assume now that $K=\mathbb{Q}_p$. However, all what is said in the following
is valid also for arbitrary $p$-adic fields.

It is  well known  that any infinite descending chain 
\begin{align}
B_1\supseteq B_2\supseteq \dots \label{descendchain}
\end{align}
of strictly smaller disks in $\mathbb{P}(\mathbb{Q}_p)$ converges to a unique point 
$$
\mathset{x}=\bigcap\limits_nB_n
$$
on the $p$-adic projective line $\mathbb{P}(\mathbb{Q}_p)$. A chain (\ref{descendchain}) defines a half\/line in 
the Bruhat-Tits tree $\mathscr{T}_{\mathbb{Q}_p}$. 

An {\em end} in a tree is an equivalence class of half\/lines, where two half\/lines are said
to be equivalent, if they differ only by finitely many  edges. 
It is a  fact that the ends of the tree $\mathscr{T}_{\mathbb{Q}_p}$ correspond bijectively to the points in $\mathbb{P}(\mathbb{Q}_p)$,
and is not too difficult to check.

\smallskip
The following subtree of the Bruhat-Tits tree is an idea of F.\ Kato \cite[\S 5.4]{FumiharuJAG2005} which turned out useful in
the study of discontinuous group actions:
\begin{dfn} \label{dendoX}
Let $X\subseteq\mathbb{P}(\mathbb{Q}_p)$ be a finite set containing $0$, $1$ and $\infty$. Then the smallest subtree $\mathscr{T}^*\langle X\rangle$
of $\mathscr{T}_{\mathbb{Q}_p}$ having $X$ as its set of ends is called the {\em projective dendrogram} for $X$.
\end{dfn}

Note that the  definition of $\mathscr{T}^*\ZR{X}$  makes sense, even if $X$ does not contain $0$, $1$ or $\infty$.

\begin{exa}\label{exa-mirror}\rm 
(1) Let $x_0,x_1\in\mathbb{P}(\mathbb{Q}_p)$ be two distinct points, and set $X=\{x_0,x_1\}$.
It defines the subtree $\mathscr{T}^{\ast}\ZR{X}$ which is a straight line:
 the geodesic in $\mathscr{T}_{\mathbb{Q}_p}$ between $x_0$ and $x_1$,
as illustrated in Figure \ref{geodesic}.
\begin{figure}[h]
$$
\xymatrix@C=50pt{
x_0&*\txt{}\ar[r]\ar[l]&x_1
}
$$
\caption{Geodesic line in $\mathscr{T}_{\mathbb{Q}_p}$.} \label{geodesic}
\end{figure}

(2) Let $X=\{x_0,x_1,x_2\}$ be a set of three mutually distinct points in $\mathbb{P}(\mathbb{Q}_p)$.
Then the subtree $\mathscr{T}^{\ast}\ZR{X}$ is a tripod, as depicted in 
Figure \ref{tripod}.
We denote by $v(x_0,x_1,x_2)$ the unique vertex of $\mathscr{T}^{\ast}\ZR{X}$ whose star  has three edges.

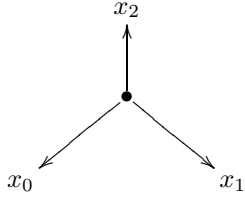
\begin{figure}[h]
$$
\xymatrix{
&x_2&\\
&*\txt{$\bullet$}\ar[u]\ar[dl]\ar[dr]&\\
x_0&&x_1
}
$$
\caption{Tripod in $\mathscr{T}_{\mathbb{Q}_p}$.} \label{tripod}
\end{figure}
\end{exa}

For a subset $X$ of $\mathbb{P}(\mathbb{Q}_p)$, define $\mathscr{T}\ZR{X}$ to be the subtree of $\mathscr{T}_{\mathbb{Q}_p}$ that is the smallest subtree among all possible subtrees containing the  vertices of the form $v(x_0,x_1,x_2)$ with $x_0,x_1,x_2\in X$.
Notice that this subtree is non-empty if and only if $X$ contains at least three points.
We call $\mathscr{T}\ZR{X}$ the {\em finite part} of the projective dendrogram $\mathscr{T}^*\ZR{X}$.
We have the obvious inclusion 
$$
\mathscr{T}\ZR{X}\longhookrightarrow\mathscr{T}^{\ast}\ZR{X}
$$
of trees.

\smallskip
It is useful to not take into account all vertices of the finite part $\mathscr{T}=\mathscr{T}\ZR{X}$
of a projective dendrogram. Consider all  paths $\gamma=[v,w]$ (without backtracking)
of maximal length in $\mathscr{T}$ whose vertices in $(v,w)$ have no  edges outside $\gamma$ emanating from them. By replacing every such path $\gamma$ of $\mathscr{T}$
by a single edge, but of equal length as $\gamma$, we obtain a so-called {\em stable tree} $\mathscr{T}^\Stab$,
whose vertices have the property that at least three edges emanate from each of them. The tree $\mathscr{T}^\Stab$
is called the {\em stabilisation} of $\mathscr{T}$.

\begin{conv}
By a (projective) dendrogram $\mathscr{T}^*=\mathscr{T}^*\ZR{X}$ we will usually mean the tree obtained by identifying the finite part $\mathscr{T}\ZR{X}$ 
 with its stabilisation $\mathscr{T}^\Stab$. 
\end{conv}

A vertex $v$ of $\mathscr{T}\ZR{X}$ is considered to be a cluster of the points corresponding to the half\/lines
in $\mathscr{T}\ZR{X}^*$ emanating from $v$. Fixing the points $0$, $1$ and $\infty$ is done for reasons of
normalisation: two points define a geodesic, three points define a unique vertex in $\mathscr{T}_{\mathbb{Q}_p}$,
and the three points $0$, $1$ and $\infty$ define the vertex $v_\mathbb{D}$ corresponding to the unit disk $\mathbb{D}$.

In this way, the usual dendrogram obtained from $\mathscr{T}^*\ZR{X}$ is 
$$
\mathscr{T}^*\ZR{X}\setminus\text{the halfline $(v_\mathbb{D},\infty)$}.
$$
A ``genuine'' dendrogram has the property that $X\subseteq \mathbb{Z}\cup\mathset{\infty}$, or, more generally, 
$\infty\neq x\in X$ has a finite expansion
$$
x=\alpha_0+\alpha_1\pi+\dots+\alpha_m\pi^m,\quad \alpha_\nu\in\mathset{0,\dots,q-1},
$$
where $\pi$ is a prime element of $O_K=\mathset{z\in K\mid \absolute{z}_K\le 1}$,
and $q$ the order of the residue field of $K$ (cf.\ \cite[\S 5]{Gouvea1993} for more details on finite field extensions of $\mathbb{Q}_p$).

\begin{Rem}
As noted in \cite{BradGfKl2007}, the task of hierarchical classification
conceptually becomes the finding of a suitable $p$-adic encoding 
which reveals the inherent hierarchical structure of data.
The reason is that the $p$-adic dendrogram $\mathscr{T}^*\langle X\rangle$
 of a given set $X\subseteq\mathbb{P}^1(\mathbb{Q}_p)$ is uniquely determined by $X$. Algorithmically, the computation of $\mathscr{T}^*\langle X\rangle$ is much
simpler than its classical counterpart \cite[\S 3.2]{Brad2007}.
\end{Rem}

\section{The space of dendrograms}

Call $M_{0,n}$ the space of all projective dendrograms for sets of cardinality $n\ge 3$.
This space is known also under the name {\em moduli space for genus $0$ curves with $n$ punctures}.
The term ``genus $0$ curve'' means nonsingular projective algebraic curve of genus $0$, i.e.\
projective line. By fixing $n$ points $x_1,\dots,x_n$ on the projective line $\mathbb{P}(\mathbb{Q}_p)$ and then
changing these points by a M\"obius transformation such that the first three are $0$, $1$, $\infty$,
we obtain a projective dendrogram. 

As moduli spaces parametrise objects up to isomorphism, and isomorphisms of punctured curves send punctures to punctures, we indeed have a moduli space $M_{0,n}$ of dendrograms by considering in each isomorphism class a normalised representative.

It is a well established fact that 
$$
M_{0,n}\cong\left(\mathbb{P}^1\setminus\mathset{0,1,\infty}\right)^{n-3}\setminus\Delta,
$$
where $\Delta$ is the fat diagonal given by $x_i=x_j$, $i\neq j$, and $\mathbb{P}^1$ is the projective line, considered as an algebraic variety 
\cite[Appendix: Lecture II]{Mumford1999}.  

One may imagine the space $M_{0,n}$ by fixing three points on $\mathbb{P}^1$ and letting the remaining $n-3$ points
vary on the projective line without collision.

In the $p$-adic setting, a family of dendrograms for $n$ points is given by a map $S\to M_{0,n}$
from some base space $S$. Each point $s\in S$ corresponds to a dendrogram, and the dendrogram
varies in some sense, as $s$ moves along $S$. 

The ``geography'' of $M_{0,n}$ is as follows: pick a dendrogram $x$ for $n$ points. Moving the points
only slightly does not change the finite part of the dendrogram.
Moving the points a little more results in changes in the lengths of the edges of $x$, but the
underlying combinatorial structure does
not change. The combinatorial tree of $x$ occupies an open subset $U$ of $M_{0,n}$. Moving points of $x$
even more results in edge contractions: by contracting one edge, $x$ moves from $U$ to a neighbouring piece $V$.
$M_{0,n}$ is covered by such disjoint open pieces, each belonging to a combinatorial tree with
$n$ ends.
 This is due to the fact that $M_{0,n}$, like many  spaces in nonarchimedean geometry,
is totally disconnected. This rather uncomfortable fact can be remedied by either resorting to
a so called {\em Grothendieck topology} or by
introducing extra points which then produce a genuine topology (e.g.\ by  considering {\em Berkovich analytic spaces} \cite{Berkovich1990}). 
This topology will be explained in the following section.

\begin{figure}[h]
$$
\begin{array}{c}
\xymatrix@=3pt{
&&\infty&&\\
 A\colon&&*\txt{$\bullet$}\ar@{-}[u]\ar@{-}[dl]\ar@{-}[ddrr]&&\\
&*\txt{$\bullet$}\ar@{-}[dl]\ar@{-}[dr]&&&\\
0&&1&&\lambda}
\end{array}
\hfill
\begin{array}{c}
\xymatrix@=3pt{
&&\infty&&\\
 B\colon&&*\txt{$\bullet$}\ar@{-}[u]\ar@{-}[dl]\ar@{-}[ddrr]&&\\
&*\txt{$\bullet$}\ar@{-}[dl]\ar@{-}[dr]&&&\\
0&&\lambda&&1}
\end{array}
\hfill
\begin{array}{c}
\xymatrix@=3pt{
&&\infty&&\\
 C\colon&&*\txt{$\bullet$}\ar@{-}[u]\ar@{-}[dl]\ar@{-}[ddrr]&&\\
&*\txt{$\bullet$}\ar@{-}[dl]\ar@{-}[dr]&&&\\
1&&\lambda&&0}
\end{array}
\hfill
\begin{array}{c}
\xymatrix@=10pt{
&\infty&\\
v\colon&*\txt{$\bullet$}\ar@{-}[u]\ar@{-}[dl]\ar@{-}[d]\ar@{-}[dr]&\\
0&1&\lambda
}
\end{array}
$$
\caption{Dendrograms representing  $M_{0,4}$.} \label{M04}
\end{figure}
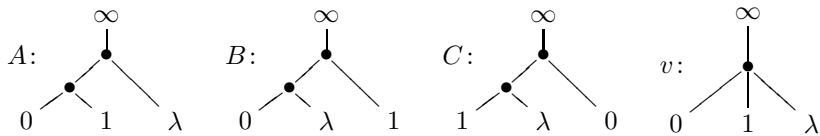

Figure \ref{M04} illustrates the dendrograms represented by 
the different parts of $M_{0,4}$:
one ``central'' region $v$
(three children) and three ``outer''  regions 
$A,B,C$ (at most two children). Any path from $A$ to $B$ or $C$ 
passes through
 $v$, as the edge has to be contracted and then blown up in a different manner.

\section{The Berkovich topology on $M_{0,n}$} \label{berkovich}

We begin with the topology on the unit disk $\mathbb{D}$ of a $p$-adic field.
The {\em classical} points of $\mathbb{D}$ are its $K$-rational points.
However, Berkovich defines in \cite{Berkovich1990} more points which correspond
to multiplicative seminorms on the algebra of power series convergent on  nonarchimedean spaces. For the unit disk this amounts to \cite[1.4.4]{Berkovich1990}: 
\begin{enumerate}
\item the classical points,
\item the disks $\mathset{x\in K\mid \absolute{x-a}_K\le r}$ 
in $\mathbb{D}$ with $r=\absolute{\epsilon}_K$, $\epsilon\in K\setminus\mathset{0}$,
\item the disks as in (2), but $0<r\neq\absolute{\epsilon}_K$ for any $\epsilon\in K$,
\item the properly descending chains $B_1\supset B_2\supset\dots$ of disks in $\mathbb{D}$ with $\bigcap B_i=\emptyset$.  
\end{enumerate}

The new points corresponding to (2), (3) or (4) are called {\em generic}, or {\em generic Berkovich points}.
This works also for the affine  line $K$, where one takes the multiplicative seminorms on the polynomial
ring $K[T]$ and obtains similarly the types (1) to (4) of points. The analogous result holds for the projective line.

The concept of generic Berkovich points via multiplicative seminorms works also in higher dimension,
and 
the result is that $p$-adic manifolds are locally contractible \cite{Berkovich1999}. In any case, by that concept, the data domain can be viewed as a 
contiunuum.

Endowing our space of dendrograms $M_{0,n}$  with the Berkovich 
topology gives us now a framework for considering continuously varying families
of dendrograms. For example, a stochastic classification of $n$ points (including $\infty$)
is nothing but a probability distribution on $M_{0,n}$, possibly with compact support.
Or the problem of adding a new datapoint to a  given classification 
$x\in M_{0,n}$ means finding a probability distribution on the fibre $\pi^{-1}(x)$,
where $\pi\colon M_{0,n+1}\to M_{0,n}$ is the map which forgets the $(n+1)$-th puncture
on the $p$-adic projective line.
A similar thing applies also to a family $S\to M_{0,n}$,
where a distribution has to be found on the fibre product
$S\times_{M_{0,n}}M_{0,n+1}$ with the map $\pi$.  

\section{Allowing collisions}
So far, our dendrograms for $n$ points can vary continuously in families, but collisions of points
are strictly excluded. In order to allow collisions, one compactifies the space $M_{0,n}$
to $\bar{M}_{0,n}$. 
 We call the points of $\partial\bar{M}_{0,n}(\mathbb{Q}_p)$ {\em stable trees of dendrograms}
 or, by abuse of language, simply {\em stable}.
 In fact, these are the so-called
{\em stable $n$-pointed trees of  projective lines} \cite{GHvdP1988}.
Such are algebraic curves $C$ which are unions of projective lines $L$
together with $n$ points $X=\mathset{x_1,\dots,x_n}\subseteq C$ and  have the defining properties:
\begin{enumerate}
\item every singular point is an ordinary double point,
\item the intersection graph of the  projective lines $L$ is a tree,
\item every projective line $L$ of which $C$ is composed contains at least three points
which are either singular points of $C$ or lie in $X$,
\item $X$ consists of regular points of $C$.
\end{enumerate}
In some sense, we can view the
points of the boundary $\partial M_{0,n}(\mathbb{Q}_p)$ as dendrograms of dendrograms. We indeed have such applications in
mind as classifications of classifications.

\begin{figure}[h]
\centering
\includegraphics[scale=.3]{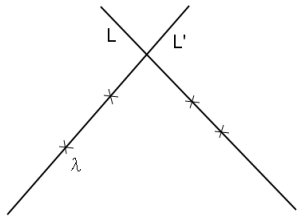}
\caption{A stable $4$-pointed tree of projective lines.}\label{stable4dendo}
\end{figure}

In order to understand what happens if a dendrogram  $x\in M_{0,n}$ moves to the boundary,
consider a dendrogram with four distinct ends $0$, $1$, $\infty$, $\lambda$, considered as
points on the projective line $L$.  The effect of $\lambda$ moving towards one of the other three
points $x$ is that, upon collision, another projective line $L'$ is formed which intersects the original
line $L$ and on which $\lambda$ and the point $x$ are again distinct. Such a configuration  corresponding
to a 
point of $\partial M_{0,4}$ is given in Figure \ref{stable4dendo}. In any case, the resulting
tree of dendrograms is indeed stable also for $n\ge 4$.

Note that the  tree with ends corresponding to a stable dendrogram does geometrically not differ from a 
projective dendrogram in $M_{0,n}$, if one forms a dendrogram for the punctures
on each of the projective lines.  The difference is
that different parts of that tree correspond to different projective lines.
This is useful for distinguishing points which are otherwise identified by collisions.

\section{Finite families of dendrograms}\label{finfam}

Assume a finite family $X$ of datasets $X_1,\dots,X_m$ each consisting of $n$ (classical) points
of the $p$-adic projective line:
$$
X_j=\mathset{x_{1j},\dots,x_{nj}},
$$
and assume at the moment that they are all different. For example, $X$ could be a 
time series $x_i(t_j)=x_{ji}$ of positions of $n$ not colliding particles
never at the same place.
Thus $X$ is the union of the $X_i$
and represents an element of $M_{0,mn}$, if we assume  $x_{11}=0$,
$x_{12}=1$ and $x_{13}=\infty$. By restricting to the points
of $X_j$ (e.g\ by taking the points at time $t_j$), we obtain a map
$$
\pi_j(X)\colon M_{0,mn}\to M_{0,n}
$$
which is the composition of the two maps

\begin{align}
(0,1,\infty,x_{14},\dots,x_{nm})&\mapsto (x_{1j},\dots,x_{nj}),\\
(x_1,\dots,x_n)&\mapsto(0,1,\infty,x_4',\dots,x'_n),
\end{align}
i.e.\ the canonical projection onto $X_j$  followed by a  M\"obius transformation $\alpha\in\PGL_2(K)$ (cf.\ Section \ref{Bruhat-Tits-tree})
which sends the first three points of $X_j$ to $0$, $1$, and $\infty$. Note that the M\"obius transformation
$\alpha=\alpha_X$ is uniquely determined by $X$ and can be easily computed.

\smallskip
If we now allow collisions of datapoints,
then we obtain a map
$$
\bar{\pi}_j(X)\colon\bar{M}_{0,mn}\to\bar{M}_{0,n},
$$
which we will not make explicit. Instead we note that if the number of distinct points of $X$
is $k$, then we have maps as before
$$
\pi_j(X)\colon M_{0,k}\to M_{0,n_j},
$$
where $n_j$ is the number of distinct points in $X_j$. The $\pi_j(X)$ are again canonical projections followed
by  M\"obius transformations, and are closely related to the maps $\bar{\pi}_j(X)$.

\smallskip 
The advantage of this moduli space approach to finite families lies in the feasibility of handling
 situations where one has a continuous family of such $X$. Moreover, the M\"obius
transformation $\alpha_X$ varies continuously with $X$.

  Again, as in Section \ref{berkovich},
one can enrich the families by probability distributions in order to obtain stochastic classifications.

\section{Hidden vertices}
\begin{dfn}
Let $\mathscr{T}^*=\mathscr{T}^*\ZR{X}$ be a projective dendrogram for $X$. A vertex $v$ of $\mathscr{T}=\mathscr{T}\ZR{X}$ is called 
{\em hidden}, if $\Star_{\mathscr{T}}(v)=\Star_{\mathscr{T}^*}(v)$.
The subgraph $\Gamma^h$ of $\mathscr{T}$ spanned by all its hidden vertices is called the {\em hidden subgraph}
of $\mathscr{T}$.
\end{dfn}

The quantity $b_0^h$, defined as the number of connected components of $\Gamma^h$,
measures how the clusters corresponding to non-hidden vertices are spread. As $\Gamma^h$ is a subgraph of
a tree, this number equals also the Euler characteristic $\chi(\Gamma^h)$.

\begin{dfn}
Let $v$ be a vertex of a graph $\Gamma$. The number $\order_\Gamma(v)=\#\Star_\Gamma(x)$
is called the {\em order} of $v$ in $\Gamma$. If $\order_\Gamma(v)=1$, then $v$ is called a {\em tip} of $\Gamma$.
\end{dfn}

By our convention, any vertex $v$ of a dendrogram has order either $1$ or greater than $2$.

\begin{thm} \label{hidvertbound}
Let $\mathscr{T}^*=\mathscr{T}^*\ZR{X}$ be a (projective) dendrogram with $\#X=n$. Then $v^h=\#\Ver(\Gamma^h)$ is bounded from above:
$$
v^h\le \frac{n}{4}-b_0^h+1.
$$
\end{thm}

\begin{proof}
{\em Case: $\Gamma^h$ connected.} If $\Gamma^h$ is connected, then either $b_0^h=1$ or $\Gamma^h=\emptyset$.
We have for the number $t^h$ of tips of $\Gamma^h$:
\begin{align}
4t^h\le n,	\label{tipbound}
\end{align}
because each tip $v$ in $\Gamma^h$ must have at least two edges in $\mathscr{T}\setminus\Gamma^h$, and, again for reasons of order,
there must be at least two ends in $\mathscr{T}^*$ emanating from each edge in $\Star_{\mathscr{T}}(v)\setminus\Star_{\Gamma^h}(v)$. This is illustrated in Figure
\ref{hiddentip}, where $v$ is a tip in $\Gamma^h$, and $e$ the unique edge in $\Star_{\Gamma^h}(v)$.

\begin{figure}[h]
\centering
\includegraphics[scale=.3]{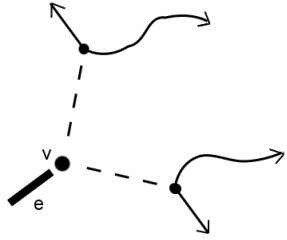}
\caption{A hidden tip in a projective dendrogram.}\label{hiddentip}
\end{figure}

Now,  the order in $\Gamma^h$ of any vertex $v$ is $0$, $1$ or $\ge 3$.
In the first case, $t^h=0$, and then 
$$
v^h=1\le\frac{n}{6}\le\frac{n}{4},
$$
where the first inequality follows in a similar way as (\ref{tipbound}).
Assume now that $\Gamma^h$ has an edge. Then
$$
v^h\le t^h\le\frac{n}{4},
$$
which is the bound in case $b_0^h=1$.

{\em General case.} In the general case, we have
$$
t^h\le \frac{n}{4}-b_0^h+1,
$$
because for each further connected component of $\Gamma^h$ there must be a path
from a tip of one
component to a tip of another in $\mathscr{T}\ZR{X}$, consisting of vertices
from which  ends of  $\mathscr{T}^*\ZR{X}$ emanate.
This proves the theorem, whether $t^h>0$ or not.
\end{proof}

\begin{cor} \label{firstcompbound}
For $X$ with $n=\#X$,
there is a bound for the number of connected components of $\Gamma^h$:
$$
b_0^h\le\frac{n+4}{8}.
$$
\end{cor}

\begin{proof}
We may assume that $\Gamma^h$ contains no edges. Then $b_0^h=v^h$, and
$$
v^h\le \frac{n}{4}-v^h+1,
$$
from which the asserted bound follows.
\end{proof}

The bound in Corollary \ref{firstcompbound} is not sharp, however. If, for example, $\Gamma^h$ is
connected and not empty, then $n$ must be at least $6$. But
$$
1<\frac{6+4}{8}
$$

\begin{thm}\label{sharpbound}
For the number of connected components of $\Gamma^h$, there is the following
sharp bound:
$$
b_0^h\le \frac{n-3}{3},
$$
where $n$ is the cardinality of $X$.
\end{thm}

\begin{figure}[h]
\centering
\begin{tabular}{cc}
\includegraphics[scale=.2]{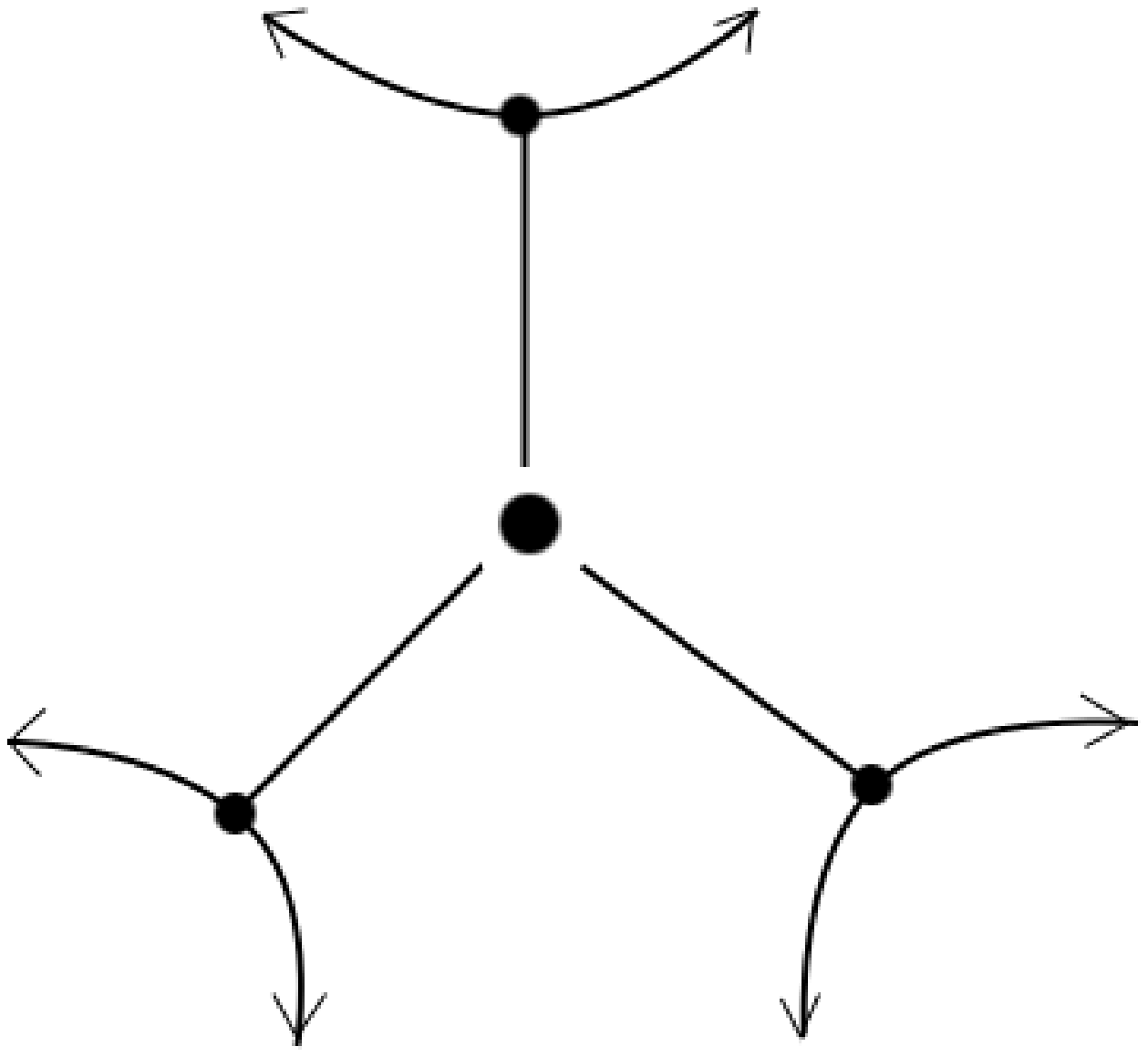}
&\includegraphics[scale=.2]{sixptdendo.eps}
\end{tabular}
\includegraphics[scale=.2]{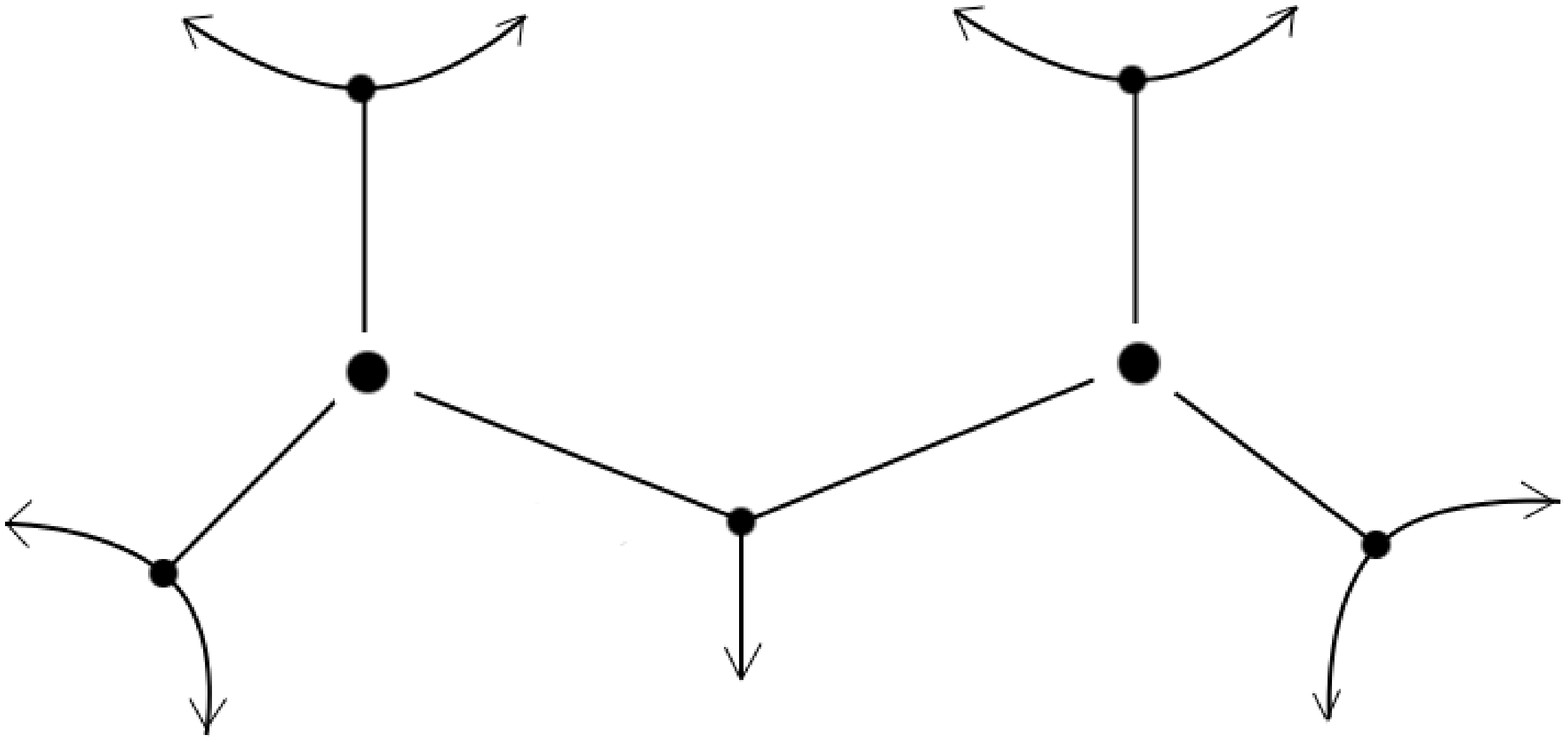}
\caption{Glueing trees along a vertex and removing three ends.}
\label{dendoglue}
\end{figure}

\begin{proof}
We may assume that $\Gamma^h$ has no edges.
By an inductive glueing of trees as in Figure \ref{dendoglue}
we obtain that for each additional connected component, one has to subtract three ends,
in order to produce a dendrogram having as few ends as possible. Thus,
$$
b_0^h\le \frac{n+3(b_0^h-1)}{6}=\frac{n-3}{6}+\frac{b_0^h}{2},
$$
from which the bound follows. Now, if $n$ is a multiple of $3$,
then $b_0^h=\frac{n-3}{3}$ by  construction. Therefore, in the general case,
$$
b_0^h=
\left
\lfloor{\frac{n-3}{3}}
\right\rfloor
$$
can be constructed. This means that the bound is sharp.
\end{proof}

\section{Conclusion}

We have given a geometric foundation for an ultrametric approach towards classification.
By extending usual dendrograms by an additional point $\infty$, they can be considered 
as points of the moduli space $M_{0,n}$ for the projective line with $n$ punctures.
The Berkovich topology allows to consider stochastic classification as giving a continuous family of dendrograms
with a probabiliy distribution on it.
The points on the boundary of $M_{0,n}$ arise from collisions of continuously evoloving datapoints and are interpreted as dendrograms of dendrograms. Time sections of
time series are given by maps $M_{0,m}\to M_{0,n}$. Finally, the topology of dendrograms is studied, resulting in bounds for the number of hidden vertices and the Euler characteristic of the
hidden graph which separates those clusters containing datapoints as maximal subclusters. 
The consequence of using $p$-adic methods is the shift of focus from
imposing a hierarchic structure on data to finding a $p$-adic encoding
which reveals the inherent hierarchies.

\section*{Acknowledgements}

The author is supported by the Deutsche Forschungsgemeinschaft (DFG) in the research project BR 3513/1-1 ``Dynamische Geb\"audebestandsklassifikation'', 
and wishes to express his gratitude for Prof.\ Dr.\ Niklaus Kohler for his 
interest in classification, and to Martin Behnisch for  drawing the author's attention 
first to \cite{Bock1974}, where he 
learned about ultrametrics in data analysis, and then to the Journal of 
Classification, where he found the
article \cite{Murtagh2004}. The latter, together with \cite{Murtagh2004a},  gave the  impetus of
writing the present article, and the unknown referees helped to improve its exposition.  
Special thanks to His Excellence Bishop-Vicar Sofian Bra\c{s}oveanul for letting the author use his office in Munich
in order to type  a substantial part of this article.

\medskip\noindent
{\sc Universit\"at Karlsruhe, Institut f\"ur Industrielle Bauproduktion, Fa\-kul\-t\"at f\"ur Architektur, Englerstr.~7, D-76128 Karlsruhe, Germany}\\
E-mail: {\tt bradley@ifib.uni-karlsruhe.de}


\end{document}